\documentclass[twocolumn,switch]{article}

\usepackage{preprint}
\usepackage[utf8]{inputenc}
\usepackage[T1]{fontenc}
\usepackage{cite}
\usepackage{amsmath,amssymb,amsfonts}
\usepackage{booktabs}
\usepackage{tabularx}
\usepackage{multirow}
\usepackage{siunitx}
\usepackage{graphicx}
\usepackage{textcomp}
\usepackage{xcolor}
\usepackage{microtype}
\usepackage{url}
\usepackage{hyperref}
\usepackage{float}
\usepackage{placeins}
\usepackage{dblfloatfix}
\usepackage{bm}

\graphicspath{{figures/}}
\definecolor{citeblue}{HTML}{1796D2}

\renewcommand{\preprintstatus}{IEEE Access paper, Publication date June 11, 2026}

\AddToHook{shipout/foreground}{%
  \begin{tikzpicture}[remember picture,overlay]
    \ifnum\value{page}=1
      \node[anchor=south, text=gray, font=\footnotesize] at ([yshift=22pt]current page.south) {Corresponding author: \texttt{jonghyuk@kookmin.ac.kr}};
    \fi
    \ifnum\value{page}=2
      \node[anchor=south, text=gray, font=\footnotesize] at ([yshift=22pt]current page.south) {Code and materials: \texttt{https://github.com/rashiedomar/CoSA}};
    \fi
  \end{tikzpicture}%
}

\hypersetup{
  pdftitle={CoSA: Correlation-Guided Change Attention with Learnable Residual Gating for Remote Sensing Change Detection},
  pdfauthor={Abdirashid Omar and Jonghyuk Park},
  pdfkeywords={change detection, remote sensing, Siamese network, decoder refinement, attention, correlation},
  colorlinks=true,
  linkcolor=black,
  urlcolor=black,
  citecolor=citeblue
}

\newcommand{\cosa}{CoSA}
\newcommand{\Fonec}{$\mathrm{F1}_c$}
\newcommand{\Iouc}{$\mathrm{IoU}_c$}
\newcommand{\PrecC}{$\mathrm{Prec}_c$}
\newcommand{\RecC}{$\mathrm{Rec}_c$}

\title{\cosa{}: Correlation-Guided Change Attention with Learnable Residual Gating for Remote Sensing Change Detection}

\author{
  Abdirashid Omar and Jonghyuk Park \\
  Department of Data Science, Graduate School of Kookmin University \\
  Seoul 02707, Republic of Korea
}

\begin{document}

\twocolumn[
  \begin{@twocolumnfalse}

\maketitle

\begin{abstract}
Remote sensing change detection (CD) from bi-temporal imagery is critical for applications such as urban monitoring, disaster assessment, and environmental management, yet robust localization remains challenging under sparse changes, noisy labels, and appearance variations.
In this paper, we propose Context Sampling Attention (CoSA), a lightweight decoder-side refinement module that explicitly leverages bi-temporal feature correlation as a control signal for adaptive change-aware feature enhancement. This differs from conventional attention mechanisms that rely on implicit feature weighting without explicit temporal control. In the implemented FC-Siam setting, CoSA computes normalized same-location cross-correlation between paired decoder features, converts low correlation into a change gate, and injects the resulting gated residual at native 1/8 and 1/16 feature scales through learnable residual scaling. This design enables effective discrimination between stable and ambiguous regions without relying on computationally expensive global attention.

Extensive experiments on four benchmark datasets (LEVIR-CD, S2Looking, DSIFN, and CLCD) demonstrate consistent improvements over strong baselines, achieving 1.5--2.6\% gains in changed-class F1 while introducing negligible parameter overhead. Ablation studies confirm that multiscale placement and learnable residual gating are both important for peak performance.

These results indicate that CoSA establishes a practical and effective refinement paradigm for enhancing temporal discriminability in Siamese change detection frameworks.

\end{abstract}

\vspace{0.35cm}

  \end{@twocolumnfalse}
]

\section{Introduction}\label{sec:introduction}
Remote sensing change detection (CD) aims to estimate a pixel-wise change mask from a co-registered bi-temporal image pair $(I_1, I_2)$ acquired over the same area at two times. The task is central to urban monitoring, disaster response, land-use analysis, and environmental management \cite{aisha2023,jungeun2025,seda2022,kyaw2024,r2024,j2026}. In practice, however, the desired semantic changes are often sparse and localized, whereas nuisance variation from illumination shifts, seasonal appearance changes, registration error, and boundary ambiguity can be large \cite{devansh2025,zainoolabadien2020,k2024,sai2026}. A useful CD model must therefore suppress spurious temporal differences while still recovering weak but genuine change regions.

Recent progress has been driven by Siamese convolutional networks, attention-enhanced decoders, and transformer-based temporal fusion models \cite{daudt2018fully,chen2020stanet,chen2022bit,bandara2022changeformer}. These architectures improve representation quality, but they also expose a practical tradeoff. Simple Siamese differencing pipelines are efficient and easy to deploy, yet they can miss weak changes or overreact to nuisance variation. Stronger global-attention models improve temporal interaction, but they often do so at higher computational cost and with greater architectural disruption \cite{yan2024,shanshan2024,wei2024,devansh2025}. This motivates a more targeted question for this paper: can a lightweight plug-in decoder module improve temporal discriminability without replacing the backbone or introducing a heavy global interaction block?

\begin{figure}[t]
  \centering
  \includegraphics[width=\columnwidth]{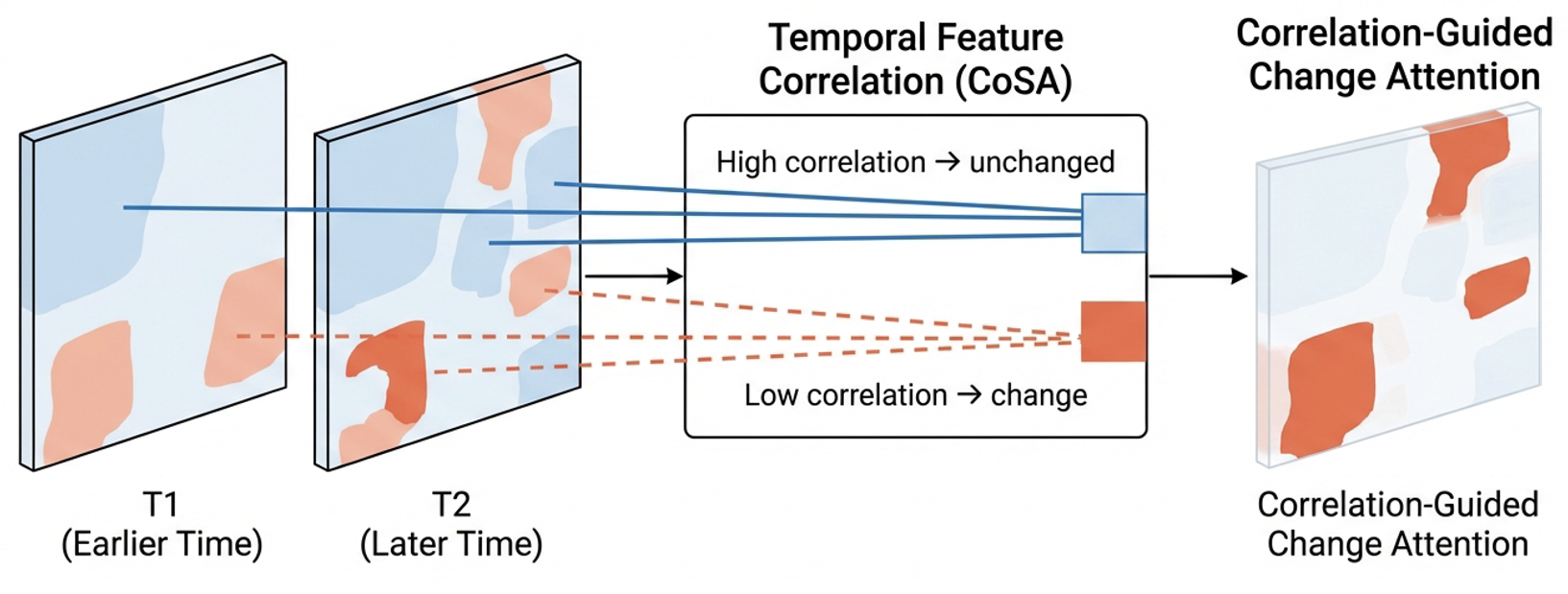}
  \caption{Conceptual overview of \cosa{} in the introduction stage. Given bi-temporal features from \(T1\) and \(T2\), \cosa{} interprets high feature correlation as unchanged context and low correlation as potential change, then applies correlation-guided change attention for refined prediction.}
  \label{fig:intro_cosa_overview}
\end{figure}

To address this problem, we propose Context Sampling Attention (\cosa{}), a decoder-side refinement module that converts bi-temporal feature correlation into an explicit change gate. In the controlled FC-Siam setting studied here, this refinement is implemented through pointwise same-location correlation at native decoder scales rather than through a separate neighborhood-sampling operator. The backbone first produces paired encoder features and absolute difference features at multiple decoder stages. \cosa{} then computes normalized cross-correlation between the paired features, transforms low correlation into a high change-response gate, and injects the gated residual into the baseline difference features through a learnable scalar. In our multiscale configuration, the same refinement is applied independently at the native 1/8 and 1/16 feature resolutions already produced by the backbone, so \cosa{} reuses existing multiresolution features instead of constructing a separate image pyramid or replacing the prediction head. Figure~\ref{fig:intro_cosa_overview} illustrates this intuition by contrasting stable high-correlation regions with ambiguous low-correlation regions that benefit from refinement.

We evaluate \cosa{} on four benchmarks with complementary difficulty profiles: LEVIR-CD, S2Looking, DSIFN, and CLCD. Across this suite, \cosa{} improves changed-class F1 by roughly 1.5--2.6 points over the controlled baseline while preserving the plug-in design. The gains are dataset dependent: some settings benefit mainly from recall recovery, others from stronger false-positive suppression, and near-neutral behavior is possible when the backbone already performs strong temporal fusion. This pattern is important because it suggests that \cosa{} is a targeted refinement mechanism rather than a universal replacement for richer temporal interaction models.

In summary, this paper makes four contributions. First, it formulates a lightweight decoder refinement strategy that uses bi-temporal correlation as an explicit control signal for change gating. Second, it instantiates this strategy as a plug-in residual module that can be inserted into existing Siamese encoder--decoder pipelines without modifying the encoder or prediction head. Third, it demonstrates consistent controlled gains on four benchmarks together with ablations that isolate the role of multiscale placement and learnable residual scaling. Fourth, it analyzes when the module helps, when it is near-neutral, and what computational overhead it introduces in the main controlled setting.

\section{Related Work}\label{sec:related_work}
Remote sensing change detection has developed from classical difference-image analysis toward deep neural architectures with increasingly explicit temporal interaction. This section briefly reviews that progression and positions \cosa{} relative to both classical and modern families.

\subsection{Classical Change Detection}
Before deep learning, remote sensing CD was commonly approached through image differencing, image ratioing, change vector analysis, principal-components-style transforms, thresholding, and post-classification comparison \cite{singh1989,lu2004,bruzzone2000}. These methods remain important because they define the core problem formulation: given two registered observations, derive a change-sensitive representation and then separate changed from unchanged regions. Their strengths are simplicity, interpretability, and low computational cost. Their limitations are equally well known: sensitivity to radiometric variation, dependence on threshold selection, and limited ability to model complex spatial context or semantic ambiguity \cite{singh1989,lu2004}. Modern learning-based CD systems inherit this same problem structure, but replace handcrafted representations and thresholds with trainable feature extractors and discriminative decoders.

\subsection{CNN-Based Change Detection}
Foundational deep CD systems are largely Siamese CNN encoder--decoder models that process the two dates with shared weights and localize change through feature differencing or late fusion \cite{daudt2018fully,ronneberger2015unet,hongruixuan2019}. This design remains attractive because it is simple, efficient, and well aligned with pixel-wise CD supervision. Subsequent CNN variants improved decoder fusion and local representation quality through denser skip connections, deeper supervision, or lightweight multi-branch refinement, as seen in DSIFN \cite{zhang2020deeply}, SNUNet-CD \cite{fang2021snunetcd}, TinyCD \cite{codegoni2022tinycd}, LightCDNet \cite{10214556}, DASNet \cite{jie2020}, adaptive Siamese fusion \cite{yunzuo2025}, asymmetric semantic CD networks \cite{kunping2022}, and multitask Siamese designs \cite{xibing2025}.

These CNN-based methods establish strong practical baselines, but their temporal interaction is often dominated by fixed differencing or local convolutional fusion. As a result, false-positive suppression and weak-change recovery can remain difficult when temporal nuisance variation is strong. This motivates plug-in refinements that preserve the efficiency of Siamese CNNs while improving how temporal evidence is used inside the decoder.

\subsection{Attention-Based Methods}
Attention modules are a common extension to Siamese CNNs because they selectively reweight features that are likely to be useful for CD. STANet \cite{chen2020stanet} introduced spatial--temporal attention for remote sensing CD, while HANet \cite{10093022} and SemiSANet \cite{chengzhen2022} explored hierarchical and graph-attentive variants. Other works refine difference features with explicit attention branches, including AFDE-Net \cite{s2023}, ADDEDNet \cite{junheng2025}, and several multi-scale architectures such as three-branch attention networks \cite{yan2024}, MDANet \cite{shanshan2024}, M2F2Net \cite{binhao2025}, wavelet-based aggregation \cite{jiangwei2024}, and lightweight multi-spectral attention models \cite{layth2025}.

These methods improve feature selection and often outperform plain CNN baselines, especially when cross-scale reasoning is important. Their limitation, however, is usually one of control: generic attention learns what to emphasize, but it does not necessarily expose an explicit temporal variable that tells the decoder where change evidence is weak or strong. \cosa{} is closer to a correlation-conditioned gate than to a generic attention branch, because its refinement strength is driven directly by bi-temporal feature agreement.

\subsection{Transformer-Based Approaches}
Transformer-based CD models push temporal interaction further by using self-attention to capture long-range dependencies across the bi-temporal pair. Representative examples include BIT \cite{chen2022bit}, ChangeFormer \cite{bandara2022changeformer}, transformer-based Siamese CD \cite{w2022}, STransUNet \cite{jianghua2022}, multitask CNN--Transformer semantic CD \cite{wei2024}, and STeInFormer \cite{xiaowen2024}. More recent work has explored stronger pretraining and larger representation models, including bitemporal foundation-model integration \cite{chen2023time}, foundation-model-based remote sensing CD \cite{10438490}, and new large-scale evaluation settings such as JL1-CD \cite{liu2025jl1}. Recent reviews also show that transformer-style global fusion is now a major axis of CD research \cite{devansh2025,kamalakar2025}.

The trade-off is practical. Global attention improves contextual modeling, but it also increases memory and computational cost as image resolution grows, and many transformer solutions require backbone-level redesign rather than decoder-side insertion. Strong temporal fusion may also reduce the marginal value of later refinement, which is consistent with our near-neutral BIT result. \cosa{} therefore targets a different operating point: it keeps the backbone intact and applies a lightweight correlation-conditioned residual gate only at selected decoder scales.

\subsection{Correlation-Based and Feature Interaction Methods}
A smaller but important line of work tries to move beyond plain differencing by making cross-temporal interaction more explicit. Changer argues that feature interaction is central to CD and introduces dedicated interaction modeling \cite{fang2022changer}. Change Guiding Network injects a learned change prior to steer prediction \cite{10234560}, while SFSCDNet uses semantic flow to encode structured temporal correspondence \cite{k2024}. Related formulations also widen the supervision or interaction setting, for example by reframing object change detection with single-temporal supervision \cite{zheng2021change} or by integrating richer bi-temporal features through large pre-trained models \cite{chen2023time}.

These methods are closest to \cosa{} in spirit because they treat temporal correspondence as something to model directly rather than as a byproduct of subtraction alone. However, they typically introduce richer interaction modules, priors, or larger fusion blocks. By contrast, \cosa{} uses a simpler pointwise cross-correlation signal to build a change gate that modulates existing difference features, and in the multiscale variant it repeats this operation independently at two backbone resolutions.

\subsection{Positioning \cosa{}}
\cosa{} is positioned between heavy backbone replacement and generic add-on attention. Relative to CNN baselines \cite{daudt2018fully,hongruixuan2019}, it injects stronger temporal control without changing the encoder or prediction head. Relative to attention-based refinements \cite{chen2020stanet,10093022,yan2024}, it uses bi-temporal correlation explicitly rather than only learning implicit reweighting. Relative to transformer-style models \cite{chen2022bit,bandara2022changeformer,wei2024}, it avoids global all-pairs interaction and remains plug-in compatible. Relative to broader interaction-based designs \cite{fang2022changer,10234560,k2024}, it emphasizes a lightweight formulation: pointwise normalized cross-correlation, multiscale decoder placement, and learnable residual scaling.

That combination is the main distinction of \cosa{}. The module is designed for targeted decoder refinement rather than wholesale architectural replacement, and the ablation later in the paper shows that the strongest gains require both multiscale placement and learnable residual gating. Recent hybrid systems for specialized CD settings, such as SiamMask-ICDNet for island-building monitoring \cite{lebao2025}, further illustrate a broader trend: many gains come from scenario-specific redesign rather than from a general refinement block. \cosa{} instead targets the unresolved design gap of an explicitly correlation-conditioned, lightweight decoder refinement that remains deployable across heterogeneous Siamese backbones.

\section{Method}\label{sec:method}
We present Context Sampling Attention (\cosa{}) as a lightweight decoder-side refinement module for Siamese change detection pipelines. The design goal is practical: improve temporal discriminability without replacing the encoder backbone or introducing a heavy global interaction stage. In the controlled FC-Siam setting used throughout this paper, the implemented \cosa{} block uses pointwise same-location correlation at native decoder scales rather than an explicit neighborhood-sampling operator. It acts on baseline absolute-difference features and returns refined difference features that are passed to the unchanged decoder and prediction head.

\subsection{Overview and Architecture}
Given a bi-temporal pair $(I_1, I_2)$, a shared Siamese encoder extracts paired features at multiple resolutions. At decoder scale $s$, let
\begin{equation}
F_1^{(s)}, F_2^{(s)} \in \mathbb{R}^{C_s \times H_s \times W_s},
\label{eq:feature_tensors}
\end{equation}
where $H_s$ is the feature-map height, $W_s$ is the feature-map width, and $C_s$ is the number of channels at scale $s$. The baseline FC-Siam decoder forms the absolute-difference tensor
\begin{equation}
X^{(s)} = \left|F_1^{(s)} - F_2^{(s)}\right|.
\label{eq:absolute_difference}
\end{equation}
Equation~\eqref{eq:feature_tensors} defines the paired decoder features, and Eq.~\eqref{eq:absolute_difference} gives the baseline difference representation that \cosa{} refines before downstream decoding. In the multiscale variant used in the main experiments, the refinement is applied independently at the native 1/8 and 1/16 resolutions produced by the backbone.

\begin{figure*}[!t]
  \centering
  \includegraphics[width=\textwidth]{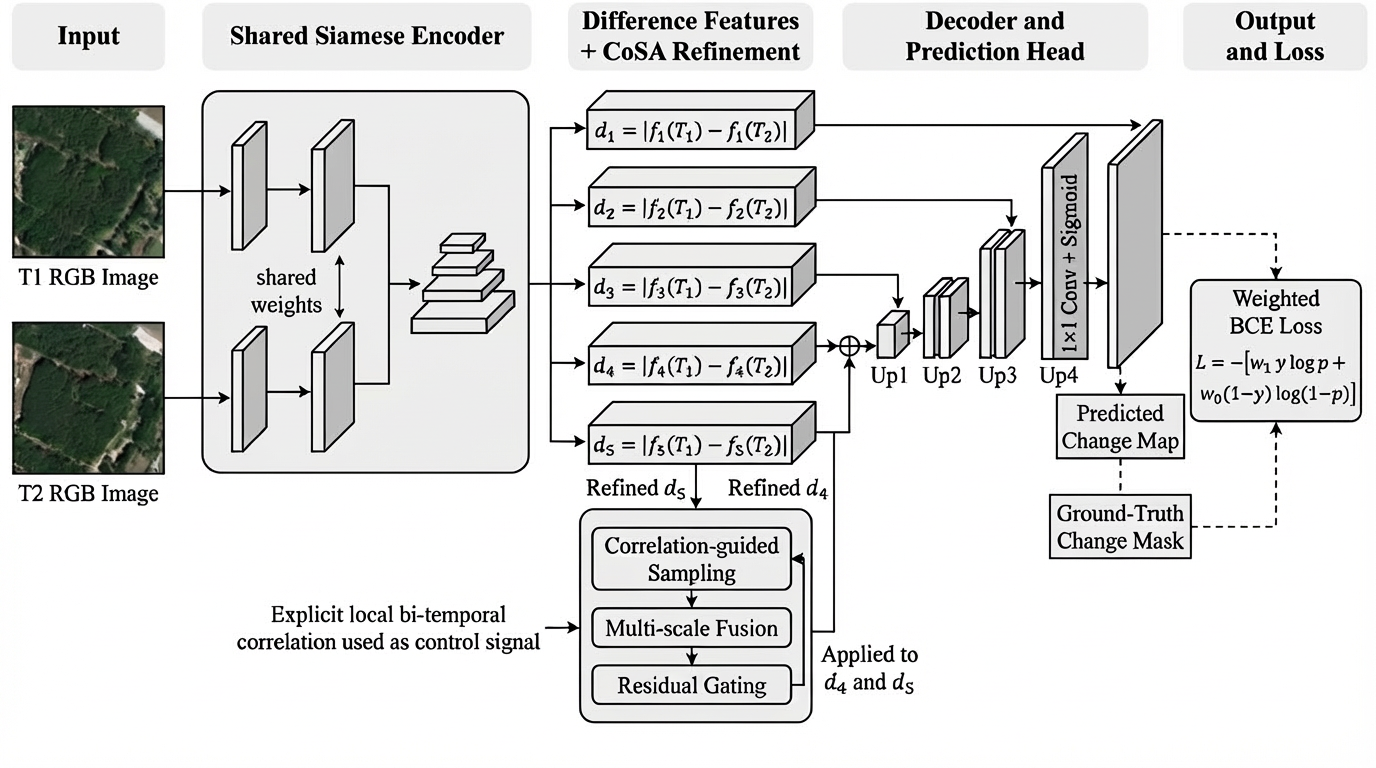}
  \caption{Full FC-Siam-style pipeline with \cosa{} inserted at selected decoder stages. The figure shows the bi-temporal inputs, shared Siamese encoder, multi-scale difference features, \cosa{} refinement applied to the 1/8 and 1/16 difference features, progressive decoding, final change-map prediction, and the weighted BCE loss used during training.}
  \label{fig:arch}
\end{figure*}

\subsection{Pointwise Bi-Temporal Correlation}
The implemented \cosa{} block uses normalized same-location cross-correlation between the paired features to quantify temporal agreement. For each scale $s$, features are $\ell_2$-normalized along the channel axis:
\begin{equation}
\bar{F}_t^{(s)}(c,x,y)=\frac{F_t^{(s)}(c,x,y)}{\sqrt{\sum_{c'=1}^{C_s} F_t^{(s)}(c',x,y)^2 + \varepsilon}},
\quad t \in \{1,2\},
\label{eq:feature_normalization}
\end{equation}
where $\varepsilon$ is a small constant for numerical stability. The scalar correlation score at pixel $(x,y)$ is then
\begin{equation}
r^{(s)}(x,y)=\sum_{c=1}^{C_s}\bar{F}_1^{(s)}(c,x,y)\,\bar{F}_2^{(s)}(c,x,y).
\label{eq:pointwise_correlation}
\end{equation}
Equation~\eqref{eq:feature_normalization} defines the channel-wise normalization, and Eq.~\eqref{eq:pointwise_correlation} defines the resulting pointwise correlation score. High $r^{(s)}(x,y)$ indicates strong temporal agreement, while low correlation indicates a potentially changed or ambiguous location.

This exact formulation is important for clarity. The controlled FC-Siam implementation used in this paper does not perform a separate neighborhood search or an explicit context-aggregation step inside \cosa{}. Correlation is computed at corresponding spatial coordinates in the paired feature maps, and refinement is driven by that pointwise temporal agreement signal.

\subsection{Correlation-to-Change Gate}
The scalar correlation map is converted into a learnable change gate. Let $T_s=\min(\texttt{topk}, C_s)$ denote the replicated gate-channel width used by the implementation. The scalar map is tiled to $T_s$ channels and passed through a $1\times1$ convolution:
\begin{equation}
\tilde{R}^{(s)} = \mathrm{Repeat}\!\left(r^{(s)}, T_s\right),
\label{eq:repeat_correlation}
\end{equation}
\begin{equation}
G^{(s)} = 1 - \sigma\!\left(W_g^{(s)} * \tilde{R}^{(s)}\right),
\label{eq:change_gate}
\end{equation}
Equation~\eqref{eq:repeat_correlation} defines the channel-repeated correlation tensor and Eq.~\eqref{eq:change_gate} defines the resulting change gate. Here $\tilde{R}^{(s)} \in \mathbb{R}^{T_s \times H_s \times W_s}$, $W_g^{(s)}$ is a learned $1\times1$ convolution that maps $T_s$ channels to one channel, $*$ denotes convolution, and $\sigma(\cdot)$ is the sigmoid function. The subtraction from 1 inverts the response so that low correlation yields a high change gate.

\subsection{Residual Refinement}
The change gate modulates the baseline difference feature through residual scaling:
\begin{equation}
\hat{X}^{(s)} = X^{(s)} + \gamma_s \left(X^{(s)} \odot G^{(s)}\right),
\label{eq:residual_refinement}
\end{equation}
Equation~\eqref{eq:residual_refinement} defines the final residual refinement step, where $\odot$ denotes element-wise multiplication and $\gamma_s$ is a learnable scalar residual gate for scale $s$. In the main configuration, $\gamma_s$ is initialized to 0, so training starts from the baseline decoder behavior and gradually learns how strongly the correlation-conditioned refinement should be injected. In the fixed-gate ablation, $\gamma_s$ is replaced by a constant scale.

This design separates temporal evidence estimation from refinement strength. The correlation term determines where temporal disagreement is high, while the residual scalar controls how aggressively that information perturbs the baseline difference feature.

\subsection{Multiscale Placement and Scaling Operations}
The multiscale CoSA variant used in the main experiments applies independent \cosa{} blocks at the 1/8 and 1/16 feature resolutions produced by the Siamese encoder--decoder backbone. These resolutions are the native backbone scales rather than new image pyramids constructed inside \cosa{}. Accordingly, the module does not introduce a separate image subsampling operator, a handcrafted neighborhood function $\mathcal{N}(x,y)$, or an additional aggregation operator $\mathrm{Agg}(\cdot)$ in the controlled FC-Siam implementation.

This distinction resolves the scaling question precisely. The only scaling operations involved are those already present in the baseline architecture: the encoder produces progressively downsampled feature maps through stride-2 stages, and the decoder upsamples them through the standard U-Net pathway. \cosa{} is attached to those native feature maps. Because the module does not create a new image pyramid, no additional low-pass anti-aliasing filter is applied inside \cosa{} itself.

\subsection{Training Objective}
\cosa{} does not introduce an auxiliary supervision branch; it is trained with the same pixel-wise changed/unchanged objective used by the underlying CD model. Let $Y \in \{0,1\}^{H \times W}$ denote the ground-truth change mask and $\hat{P} \in [0,1]^{H \times W}$ the predicted changed-class probability map. We optimize a weighted binary cross-entropy loss:
\begin{equation}
\mathcal{L}_{\text{wBCE}} = - \sum_{x,y} \Big[w_1 Y_{x,y}\log \hat{P}_{x,y}
+ w_0(1-Y_{x,y})\log(1-\hat{P}_{x,y})\Big],
\label{eq:wbce}
\end{equation}
Equation~\eqref{eq:wbce} defines the weighted training objective, where $w_1$ and $w_0$ balance the changed and unchanged classes, respectively. In our controlled experiments, baseline and \cosa{} use the same unchanged:changed weighting ratio of $1:3$, so the reported differences arise from decoder refinement rather than a modified loss function.

\subsection{Workflow and Integration Scope}
As illustrated in Figure~\ref{fig:arch}, the implemented \cosa{} block performs four steps at each selected decoder scale: compute the baseline absolute-difference feature $X^{(s)}$; estimate pointwise normalized cross-correlation from the paired features $F_1^{(s)}$ and $F_2^{(s)}$; convert low correlation into a change gate $G^{(s)}$; and inject the gated residual into $X^{(s)}$ through the learnable scalar $\gamma_s$. The encoder backbone, differencing strategy, decoder topology, and prediction head remain unchanged.

The practical defaults used in the main FC-Siam experiments are two scales (1/8 and 1/16), gate-channel width $\texttt{topk}=32$, and near-zero residual initialization for stable warm-up. Under this formulation, the main novelty of \cosa{} is not a heavyweight temporal matching module, but a lightweight correlation-conditioned residual gate that can be inserted into existing Siamese decoders with negligible parameter overhead.
Figure~\ref{fig:arch} emphasizes that this refinement is confined to the selected decoder stages and leaves the backbone and prediction head unchanged.

\section{Experiments}\label{sec:experiments}
This section defines the datasets, metrics, and evaluation protocol used to assess \cosa{} under controlled and transfer settings. Table~\ref{tab:datasets} summarizes the four benchmark datasets used throughout the section (scale, splits, and resolution).

\begin{table*}[t]
  \centering
  \caption{Dataset statistics for the four benchmarks used in this paper. Dataset sources and access links are given in the corresponding citations: LEVIR-CD \cite{chen2020stanet,levircd_dataset}, S2Looking \cite{s2looking_dataset}, DSIFN \cite{dsifn_dataset}, and CLCD \cite{clcd_dataset}.}
  \label{tab:datasets}
  \setlength{\tabcolsep}{7pt}
  \renewcommand{\arraystretch}{1.08}
  \small
  \begin{tabular*}{\textwidth}{@{\extracolsep{\fill}} l c c c c c}
    \toprule
    Dataset & Task & Image Pairs & Image Size & Change Instances & Change Pixels \\
    \midrule
    LEVIR-CD & binary & 637 & 1024$\times$1024 & 31K & 30M \\
    S2Looking & binary & 5,000 & 1024$\times$1024 & 66K & 69M \\
    DSIFN & binary & 394 & 512$\times$512 & -- & -- \\
    CLCD & binary & 560 & 512$\times$512 & -- & -- \\
    \bottomrule
  \end{tabular*}
\end{table*}

\subsection{Datasets and Evaluation Metrics}
We evaluate on four benchmarks with complementary difficulty profiles: LEVIR-CD \cite{chen2020stanet}, S2Looking \cite{s2looking_dataset}, DSIFN \cite{dsifn_dataset}, and CLCD \cite{clcd_dataset}. This selection intentionally spans clean urban building change (LEVIR-CD), sparse off-nadir scenes (S2Looking), mixed urban-rural conditions (DSIFN), and noisy agricultural boundaries (CLCD), so claims are not tied to one favorable benchmark. The statistics and sources in Table~\ref{tab:datasets} make clear that the four benchmarks differ substantially in annotation richness and scene structure.

\begin{figure*}[!t]
  \centering
  \includegraphics[width=\textwidth]{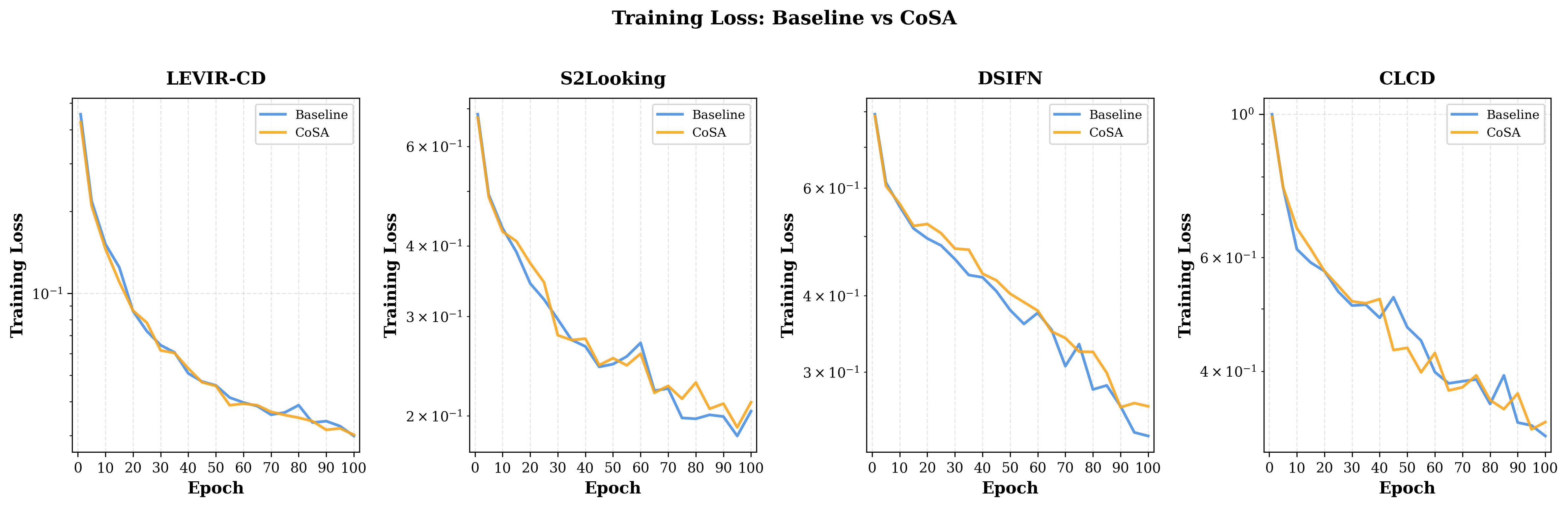}
  \caption{Training loss for the FC-Siam baseline and FC-Siam+\cosa{} across LEVIR-CD, S2Looking, DSIFN, and CLCD.}
  \label{fig:training_dynamics}
\end{figure*}

Figure~\ref{fig:training_dynamics} summarizes the training-loss trajectories for the controlled FC-Siam baseline and FC-Siam+\cosa{} runs on each of the four datasets under the shared protocol. These four datasets are sufficient for evaluating generality because, taken together, they cover the main axes on which remote sensing CD methods typically fail: clean versus noisy annotation, dense versus sparse change, stable versus off-nadir acquisition, and relatively sharp versus ambiguous boundaries. LEVIR-CD provides a relatively clean, high-quality setting with limited headroom; S2Looking stresses sparse and ambiguous changes under viewpoint variation; DSIFN represents intermediate real-world complexity; and CLCD stresses robustness to noisy boundaries and label uncertainty. A method that improves consistently across all four is therefore less likely to be benefiting from dataset cherry-picking and more likely to be addressing a transferable refinement problem.

Following standard CD reporting, we use changed-class precision, recall, F1, and IoU (\PrecC{}, \RecC{}, \Fonec{}, \Iouc{}) \cite{devansh2025,kamalakar2025}. Let $TP$, $FP$, and $FN$ denote changed-class true positives, false positives, and false negatives, respectively. The metrics are
\begin{equation}
\mathrm{Prec}_c=\frac{TP}{TP+FP},
\label{eq:precision_c}
\end{equation}
\begin{equation}
\mathrm{Rec}_c=\frac{TP}{TP+FN},
\label{eq:recall_c}
\end{equation}
\begin{equation}
\mathrm{F1}_c=\frac{2\cdot \mathrm{Prec}_c\cdot \mathrm{Rec}_c}{\mathrm{Prec}_c+\mathrm{Rec}_c},
\label{eq:f1_c}
\end{equation}
\begin{equation}
\mathrm{IoU}_c=\frac{TP}{TP+FP+FN}.
\label{eq:iou_c}
\end{equation}
Equations~\eqref{eq:precision_c}--\eqref{eq:iou_c} define the changed-class precision, recall, F1, and IoU used throughout the paper. These metrics capture false-alarm suppression (precision), missed-change recovery (recall), overall discrimination (F1), and spatial overlap quality (IoU).

\subsection{Experimental Settings and Protocol}
To isolate architectural effects, baseline and \cosa{} share identical settings: a Siamese U-Net-style decoder \cite{ronneberger2015unet} with a ResNet-18 backbone \cite{he2016resnet}, absolute differencing, weighted cross-entropy (unchanged:changed = 1:3), Adam ($\beta_1=0.9$, $\beta_2=0.999$), initial learning rate $10^{-4}$, and a ReduceLROnPlateau scheduler that monitors validation \Fonec{} and halves the learning rate when the score fails to improve for 10 epochs. Training uses batch size 8 for up to 100 epochs with early stopping (patience 15) and seed 42. We use standard augmentation (flip, rotation, and light color jitter), select checkpoints by validation \Fonec{}, and report test metrics at a 0.5 sigmoid threshold. For LEVIR-CD, the split is 445 training images, 64 validation images, and 128 test images. For S2Looking, DSIFN, and CLCD, we use fixed train/validation/test splits documented in the released configs to ensure reproducibility. Standalone STANet and BIT experiments are run under their native/custom pipelines to evaluate plug-in compatibility under model-preferred recipes; these are reported in Subsection~\ref{sec:hero_models}.
The smoother optimization behavior visible in Figure~\ref{fig:training_dynamics} is consistent with the shared protocol used for these controlled comparisons.

\subsection{Baseline vs \cosa{}}
Table~\ref{tab:fcsiam_vs_cosa} reports controlled test-set results for the FC-Siam baseline and FC-Siam+\cosa{} across all four datasets.

\begin{table*}[t]
  \centering
  \caption{Performance comparison between the FC-Siam baseline and \cosa{} on four change detection benchmarks.}
  \label{tab:fcsiam_vs_cosa}
  \setlength{\tabcolsep}{6pt}
  \renewcommand{\arraystretch}{1.08}
  \small
  \begin{tabular*}{\textwidth}{@{\extracolsep{\fill}} l l r r r r r r}
    \toprule
    Dataset & Method & \Fonec{} (\%) & \Iouc{} (\%) & \PrecC{} (\%) & \RecC{} (\%) & $\Delta \mathrm{F1}_c$ & $\Delta \mathrm{IoU}_c$ \\
    \midrule
    \textbf{LEVIR-CD} & FC-Siam baseline & 87.70 & 78.09 & 90.23 & 85.31 & --- & --- \\
    & \cosa{} & \textbf{89.45} & \textbf{80.91} & \textbf{90.39} & \textbf{88.52} & \textbf{+1.75} & \textbf{+2.82} \\
    \midrule
    \textbf{DSIFN} & FC-Siam baseline & 74.85 & 59.81 & \textbf{82.05} & 68.81 & --- & --- \\
    & \cosa{} & \textbf{76.52} & \textbf{61.98} & 68.61 & \textbf{86.50} & \textbf{+1.67} & \textbf{+2.17} \\
    \midrule
    \textbf{CLCD} & FC-Siam baseline & 46.94 & 30.67 & 34.16 & \textbf{75.04} & --- & --- \\
    & \cosa{} & \textbf{49.53} & \textbf{32.91} & \textbf{41.97} & 60.40 & \textbf{+2.59} & \textbf{+2.24} \\
    \midrule
    \textbf{S2Looking} & FC-Siam baseline & 38.00 & 23.46 & 24.93 & 79.82 & --- & --- \\
    & \cosa{} & \textbf{40.44} & \textbf{25.34} & \textbf{26.99} & \textbf{80.58} & \textbf{+2.44} & \textbf{+1.88} \\
    \bottomrule
  \end{tabular*}
\end{table*}

\begin{figure*}[!b]
  \centering
  \includegraphics[width=0.96\textwidth]{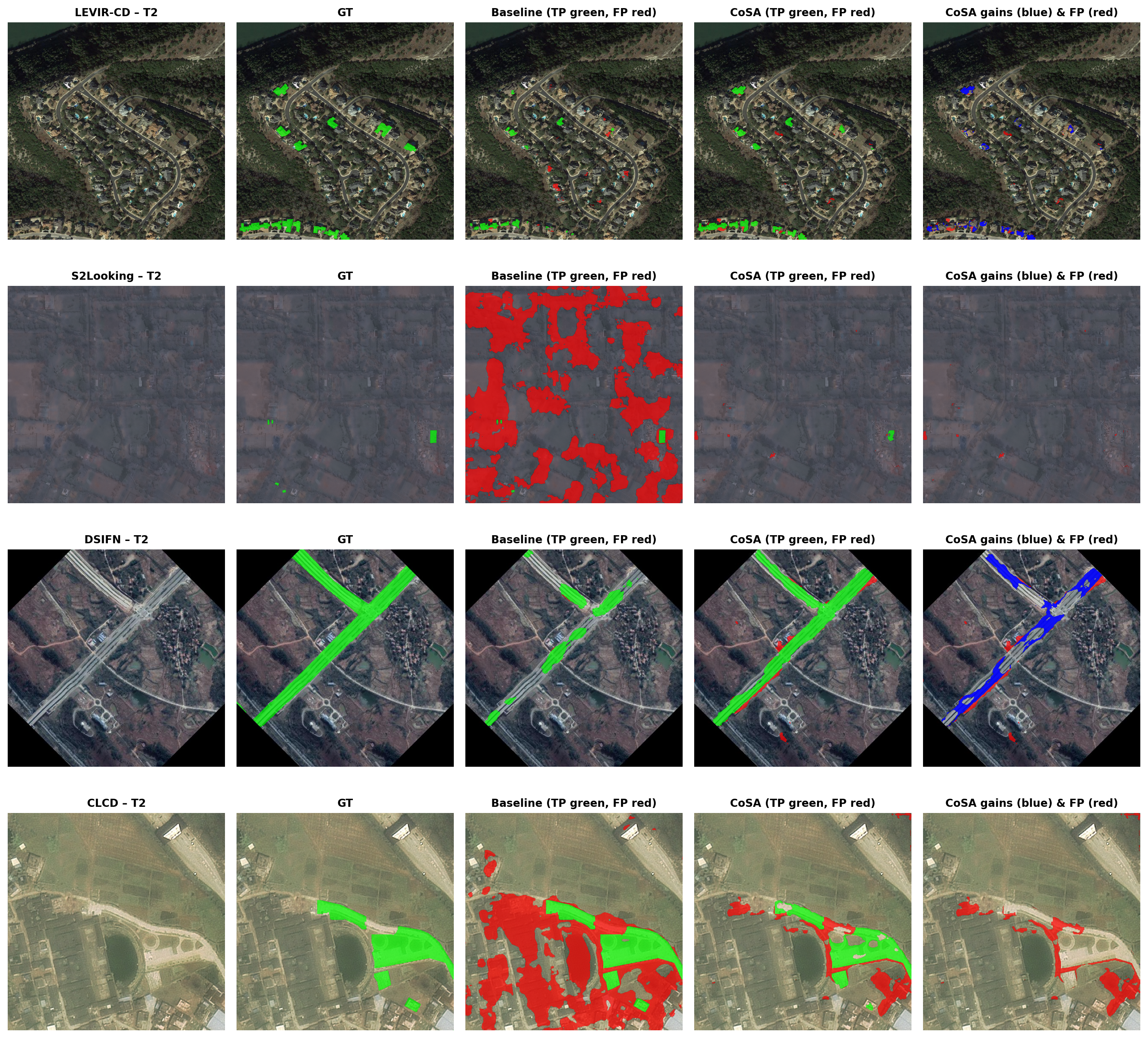}
  \caption{Qualitative baseline-vs-\cosa{} comparison across LEVIR-CD, S2Looking, DSIFN, and CLCD. Columns (left to right): T2 image, ground truth, baseline overlay (TP green, FP red), \cosa{} overlay (TP green, FP red), and \cosa{} gains (blue) with remaining FP (red).}
  \label{fig:baseline_cosa_qual_all_datasets}
\end{figure*}

\cosa{} improves \Fonec{} and \Iouc{} on all four datasets, with \Fonec{} gains from +1.67 to +2.59 and \Iouc{} gains from +1.88 to +2.82. Although the numerical gains may appear modest, they are consistent across all datasets and achieved with negligible parameter overhead. The largest gains appear on S2Looking and CLCD, where sparse changes and noisy boundaries make correlation-conditioned refinement more valuable. The gains are not uniform: on LEVIR-CD, \cosa{} preserves precision and improves recall (+3.21 points), indicating stronger missed-change recovery without clear false-alarm inflation; on CLCD, precision rises strongly (+7.81) while recall decreases, indicating conservative false-positive filtering in noisy conditions; DSIFN shows a recall-oriented shift with reduced precision, while S2Looking improves both precision and recall modestly. This behavior reflects the adaptive nature of \cosa{}, which adjusts refinement based on dataset characteristics. The training-loss curves are consistent with this pattern: \cosa{} typically converges to lower loss and stabilizes earlier across the four datasets. This dataset-dependent behavior motivates the deeper error analysis in Section~\ref{sec:analysis}.

Qualitative examples are summarized in Figure~\ref{fig:baseline_cosa_qual_all_datasets} for one strong-gain test sample per dataset under the same protocol.

Figure~\ref{fig:baseline_cosa_qual_all_datasets} makes these dataset-specific correction patterns visually concrete.
The visual trends align with Table~\ref{tab:fcsiam_vs_cosa}: reduced false positives in sparse/noisy scenes and improved coverage of weak true changes in ambiguous regions.

We isolate mechanism-level evidence in Subsection~\ref{sec:ablation}, test cross-backbone transferability in Subsection~\ref{sec:hero_models}, and analyze FP/FN transitions with patch-level consistency in Section~\ref{sec:analysis}.

\subsection{Reproducibility}
We fix training and data-loading seeds (including seed 42 for the controlled FC-Siam runs), save the best checkpoint by validation \Fonec{}, and cache per-sample predictions used for analysis tables. Paper tables and figures are generated from those logged artifacts. Full datasets and large checkpoints are omitted from the public repository for size, but released code includes configuration files, split definitions, training scripts, and evaluation commands sufficient to reproduce the reported metrics on the same public benchmarks.

Beyond the core training code, the public repository preserves the exact paper-generation path used for the controlled FC-Siam results. The released files document the train/validation/test partitions for each benchmark, the checkpoint-selection rule, the fixed 0.5 decision threshold used for changed-class reporting, and the evaluation scripts that export \PrecC{}, \RecC{}, \Fonec{}, and \Iouc{} from saved predictions. For the analysis section, the code also caches per-image confusion counts, sample-wise F1 deltas, and overlay visualizations so that the error tables and qualitative figures are regenerated from stored artifacts rather than reconstructed manually. This is especially important because the paper mixes a shared controlled recipe for FC-Siam with model-native transfer tests for STANet and BIT; the released commands separate those pipelines explicitly and preserve which results belong to the controlled protocol versus the model-preferred protocol.

\FloatBarrier

\subsection{Ablation Study}\label{sec:ablation}
\vspace{-0.7em}
\begin{figure}[H]
  \centering
  \includegraphics[width=\columnwidth]{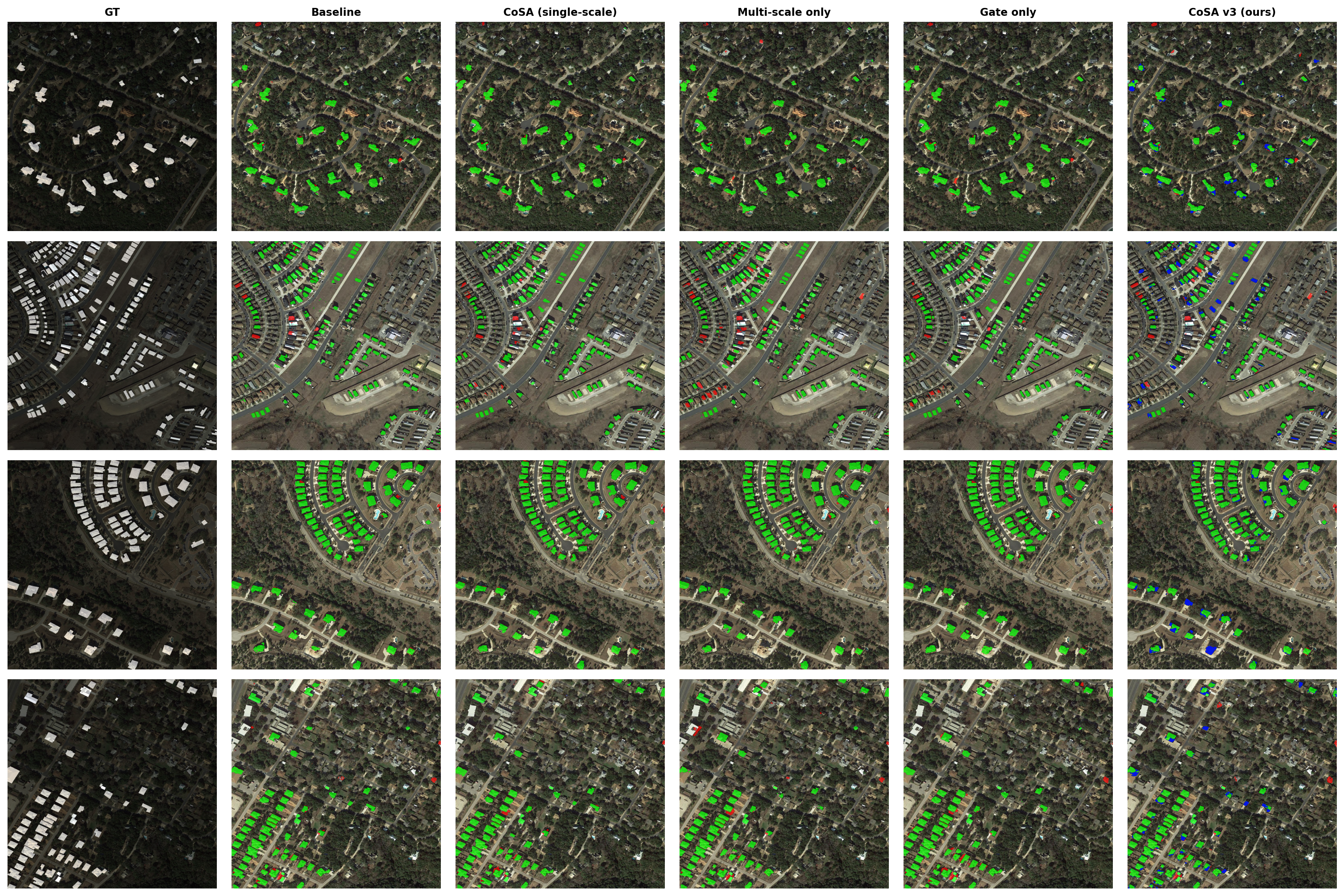}
  \caption{Qualitative ablation comparison (errors only): highlighted regions emphasize false positives and false negatives across variants.}
  \label{fig:ablation_qualitative}
\end{figure}

We conduct an ablation study over multiple variants on the same FC-Siam-diff-style backbone: baseline, attention-only, \cosa{} with multi-scale only, \cosa{} with gate only, \cosa{} with single-scale and fixed gate, alignment-first, and the full \cosa{} configuration with multi-scale refinement and learnable gating.

Table~\ref{tab:ablation} lists quantitative results for each variant on LEVIR-CD. Figure~\ref{fig:ablation_qualitative} complements Table~\ref{tab:ablation} with qualitative error maps (false positives and false negatives) for the same setting. Representative complexity metrics are summarized separately in Table~\ref{tab:efficiency_overhead}.

\begin{table*}[t]
  \centering
  \caption{Ablation study on LEVIR-CD (changed class).}
  \label{tab:ablation}
  \setlength{\tabcolsep}{6pt}
  \renewcommand{\arraystretch}{1.05}
  \small
  \begin{tabularx}{\textwidth}{X r r r r}
    \toprule
    Variant & \PrecC{} & \RecC{} & \Fonec{} & \Iouc{} \\
    \midrule
    Baseline & 90.23 & 85.31 & 87.70 & 78.09 \\
    Attention-only & 90.27 & 85.61 & 87.88 & 78.38 \\
    \cosa{} with multi-scale only & 91.35 & 83.68 & 87.35 & 77.54 \\
    \cosa{} with gate only & 93.33 & 82.67 & 87.68 & 78.06 \\
    \cosa{} with single-scale and fixed gate & \textbf{93.88} & 83.22 & 88.23 & 78.94 \\
    Alignment-first & 92.00 & 85.62 & 88.70 & 79.69 \\
    Full \cosa{} (multi-scale + learnable gate; ours) & 90.39 & \textbf{88.52} & \textbf{89.45} & \textbf{80.91} \\
    \bottomrule
  \end{tabularx}
\end{table*}

``Multi-scale'' inserts \cosa{} at two resolutions (1/8 and 1/16). ``Learnable gate'' uses a trainable residual scale $\gamma$ (Section~\ref{sec:method}) initialized near zero; fixed-scale uses a constant residual factor. ``Gate only'' disables multi-scale placement and keeps only residual gating at the bottleneck. ``Alignment-first'' estimates local displacement and warps features before differencing. The full \cosa{} configuration (multi-scale + learnable residual scale) gives the best F1/IoU balance among tested variants. Relative to baseline, \cosa{} improves recall (+3.21 points) with near-stable precision (+0.16 points), yielding +1.75 \Fonec{} and +2.82 \Iouc{}. The single-scale fixed variant improves precision but reduces recall, indicating that correlation-conditioned refinement without adaptive residual control can become overly conservative. The alignment-first variant improves over baseline but remains below full \cosa{}, suggesting that the implemented pointwise correlation gate is more effective than geometric pre-alignment alone in this setup \cite{shanshan2024,yan2024}. Table~\ref{tab:ablation} therefore provides the experimental evidence for the method-level claim: multiscale placement alone or residual gating alone is insufficient for peak performance, and the strongest gain appears only when both multiscale placement and adaptive residual control are used together. Figure~\ref{fig:ablation_qualitative} illustrates where partial variants fail: the full module suppresses spurious detections while recovering missing change regions more consistently than partial variants.

\subsection{Standalone Hero Models}\label{sec:hero_models}
We report standalone ``hero'' experiments for STANet \cite{chen2020stanet} and BIT \cite{chen2022bit} under their native/custom pipelines. The goal is to test transferability of \cosa{} when inserted into strong non-baseline backbones.
Table~\ref{tab:hero_models} reports the changed-class results for these two backbones and shows that \cosa{} delivers a large gain on STANet but only a marginal change on BIT. Their representative computational profiles are summarized separately in Table~\ref{tab:efficiency_overhead}.

\begin{table*}[t]
  \centering
  \caption{Standalone hero models under their native pipelines (changed class).}
  \label{tab:hero_models}
  \setlength{\tabcolsep}{7pt}
  \renewcommand{\arraystretch}{1.05}
  \small
  \begin{tabular*}{\textwidth}{@{\extracolsep{\fill}} l l r r r r}
    \toprule
    Model & Variant & \PrecC{} & \RecC{} & \Fonec{} & \Iouc{} \\
    \midrule
    STANet & Baseline & 70.31 & \textbf{94.71} & 80.71 & 67.66 \\
    STANet & + \cosa{} & \textbf{87.11} & 85.19 & \textbf{86.14} & \textbf{75.66} \\
    BIT & Baseline & 91.37 & \textbf{91.08} & 91.22 & 83.87 \\
    BIT & + \cosa{} & \textbf{92.38} & 90.17 & \textbf{91.26} & \textbf{83.93} \\
    \bottomrule
  \end{tabular*}
\end{table*}

The two hero models exhibit clearly different responses. On STANet, \cosa{} provides a large gain (\Fonec{}: 80.71 \textrightarrow 86.14; \Iouc{}: 67.66 \textrightarrow 75.66) with a strong precision increase, indicating effective suppression of over-detection. On BIT, the effect is near-neutral (\Fonec{}: 91.22 \textrightarrow 91.26), consistent with partial redundancy between explicit correlation gating and BIT's transformer-based temporal interaction. Standalone evidence therefore matches the broader pattern in this paper: \cosa{} is most beneficial when it complements, rather than duplicates, the temporal interaction already encoded by the backbone \cite{w2022,wei2024}.

\section{Analysis}\label{sec:analysis}
This section moves beyond aggregate scores to analyze \emph{how} \cosa{} changes errors, \emph{where} gains occur at sample level, and \emph{why} behavior differs across datasets and backbones.

\subsection{Error Correction Mechanics: FP/FN Transitions}
To expose correction behavior, we compare baseline and \cosa{} error counts on the same test split. Let $\mathrm{FP}_{\text{base}}$ and $\mathrm{FP}_{\text{cosa}}$ denote the changed-class false-positive pixel counts before and after inserting \cosa{}, and define $\mathrm{FN}_{\text{base}}$ and $\mathrm{FN}_{\text{cosa}}$ analogously for false negatives. We report percentage reduction as
\begin{equation}
\mathrm{FP\ Red.}=\frac{\mathrm{FP}_{\text{base}}-\mathrm{FP}_{\text{cosa}}}{\mathrm{FP}_{\text{base}}}\times 100,
\label{eq:fp_reduction}
\end{equation}
\begin{equation}
\mathrm{FN\ Red.}=\frac{\mathrm{FN}_{\text{base}}-\mathrm{FN}_{\text{cosa}}}{\mathrm{FN}_{\text{base}}}\times 100.
\label{eq:fn_reduction}
\end{equation}
Equations~\eqref{eq:fp_reduction} and \eqref{eq:fn_reduction} define the error-reduction statistics reported in Table~\ref{tab:error_mechanics}. Positive values indicate fewer errors with \cosa{}; negative values indicate error increase.

\begin{table}[H]
  \centering
  \caption{Error correction mechanics for \cosa{} on the LEVIR-CD test set across the controlled FC-Siam model and the standalone STANet and BIT settings. Entries report percent reduction in changed-class FP/FN counts computed as $(\mathrm{baseline}-\mathrm{\cosa{}})/\mathrm{baseline}\times100$ on cached per-sample confusion counts. Positive values mean \cosa{} reduces that error count; negative values mean \cosa{} increases it.}
  \label{tab:error_mechanics}
  \setlength{\tabcolsep}{6pt}
  \renewcommand{\arraystretch}{1.05}
  \small
  \begin{tabular*}{\columnwidth}{@{\extracolsep{\fill}} l c c}
    \toprule
    Model family & FP reduction & FN reduction \\
    \midrule
    CNN (FC-Siam) & -30.3\% & \textbf{+23.7\%} \\
    Attention (STANet) & \textbf{+68.5\%} & -179.8\% \\
    Transformer (BIT) & +13.6\% & -10.2\% \\
    \bottomrule
  \end{tabular*}
\end{table}

\noindent\textbf{Interpretation.} Table~\ref{tab:error_mechanics} shows that the correction strategy is backbone-dependent. In the controlled CNN setting (FC-Siam), \cosa{} is recall-oriented: FN decreases while FP rises. In STANet, behavior flips to precision-oriented filtering: FP drops strongly while FN increases. In BIT, shifts are comparatively small, consistent with partial redundancy when temporal interaction is already strong \cite{chen2022bit,w2022,wei2024}. This confirms that \cosa{} is adaptive rather than uniformly precision- or recall-biased.

\subsection{Patch-Level Consistency and Sample-Wise Distribution}
For each sample $i$, we compute:
\begin{equation}
\Delta \mathrm{F1}_c^{(i)} = \mathrm{F1}_{c,\text{cosa}}^{(i)} - \mathrm{F1}_{c,\text{base}}^{(i)},
\label{eq:delta_f1_sample}
\end{equation}
where $\mathrm{F1}_{c,\text{base}}^{(i)}$ and $\mathrm{F1}_{c,\text{cosa}}^{(i)}$ are the changed-class sample-wise F1 scores before and after inserting \cosa{}. Equation~\eqref{eq:delta_f1_sample} is then used to group samples into four buckets: significant gain $(\Delta \mathrm{F1}_c^{(i)} > 10)$, moderate gain $(2 \le \Delta \mathrm{F1}_c^{(i)} \le 10)$, neutral $(-1 \le \Delta \mathrm{F1}_c^{(i)} < 2)$, and degradation $(\Delta \mathrm{F1}_c^{(i)} < -1)$.

\begin{table}[t]
  \centering
  \caption{Patch-level consistency of \cosa{} on the LEVIR-CD test set. Each cell reports the percentage of samples in a category and the average $\Delta$\Fonec{} (points) within that category.}
  \label{tab:consistency}
  \setlength{\tabcolsep}{4pt}
  \renewcommand{\arraystretch}{1.05}
  \scriptsize
  \begin{tabular*}{\columnwidth}{@{\extracolsep{\fill}} l c c c}
    \toprule
    Category & FC-Siam & STANet & BIT \\
    \midrule
    Significant gain & \textbf{6.25\%}, +16.97 & 5.42\%, +19.13 & 2.10\%, \textbf{+90.50} \\
    Moderate gain & 21.09\%, +5.00 & \textbf{21.83\%}, \textbf{+5.52} & 1.42\%, +4.58 \\
    Neutral & 58.59\%, \textbf{+0.58} & 63.67\%, +0.05 & \textbf{89.31\%}, +0.04 \\
    Degradation & \textbf{14.06\%}, \textbf{-5.78} & 9.08\%, -22.97 & 7.18\%, -10.87 \\
    \bottomrule
  \end{tabular*}
\end{table}

\noindent\textbf{Interpretation.} Most samples are neutral across all families (especially BIT), indicating that refinement is largely non-interfering. For FC-Siam and STANet, improvement buckets (moderate + significant) exceed degradation frequency, which explains positive aggregate gains. BIT remains predominantly neutral, consistent with its near-ceiling baseline and marginal overall gain in standalone tests.

\noindent\textbf{When \cosa{} helps.} Strongest improvements are concentrated in FN-heavy and boundary-ambiguous samples, where pointwise bi-temporal correlation cues recover coherent changed regions missed by baseline predictions.

\noindent\textbf{When \cosa{} can hurt.} Degradation is concentrated in ambiguous or noisy-label regions and weak-texture areas, where correlation structure is less reliable.

\begin{table*}[!t]
  \centering
  \caption{Representative complexity comparison on LEVIR-CD. Parameters and GFLOPs are reported per model, while latency, FPS, and peak GPU memory are measured on an RTX A6000 under a shared paired-input forward-pass benchmark (batch size 1, $256\times256$, evaluation mode).}
  \label{tab:efficiency_overhead}
  \setlength{\tabcolsep}{4pt}
  \renewcommand{\arraystretch}{1.05}
  \scriptsize
  \begin{tabular*}{\textwidth}{@{\extracolsep{\fill}} l r r r r r}
    \toprule
    Method & Params (M) & GFLOPs@256 & Latency (ms) & FPS & Peak Mem. (MB) \\
    \midrule
    FC-Siam baseline & 33.227841 & 137.5690 & 6.82 & 146.59 & 531.62 \\
    FC-Siam + \cosa{} & 33.227907 & 137.5691 & 6.72 & 148.81 & 661.32 \\
    STANet baseline & 16.897041 & 26.2054 & 8.50 & 117.58 & 1016.94 \\
    STANet + \cosa{} & 17.016917 & 29.2512 & 9.23 & 108.33 & 1017.42 \\
    BIT baseline & 11.943754 & 17.5045 & 15.26 & 65.55 & 519.68 \\
    BIT + \cosa{} & 11.943787 & 17.5048 & 15.29 & 65.39 & 521.20 \\
    \bottomrule
  \end{tabular*}
\end{table*}

\subsection{Cross-Dataset Adaptive Behavior}
Table~\ref{tab:consistency} provides the sample-wise backdrop for the cross-dataset trends discussed next. Controlled results in Table~\ref{tab:fcsiam_vs_cosa} show that \cosa{} improves \Fonec{} on all four datasets, but via different precision/recall routes. LEVIR-CD benefits mainly from recall recovery with stable precision; S2Looking improves both precision and recall; DSIFN shifts strongly toward recall; and CLCD shifts toward precision filtering under noisier boundaries. This pattern supports the method claim from Section~\ref{sec:method}: learnable residual gating enables dataset-dependent calibration rather than fixed global behavior.

\subsection{Computational Cost and Efficiency Discussion}
\cosa{} is implemented as a pointwise correlation-conditioned residual gate, so its cost is lighter than the neighborhood-aggregation formulation used in the previous draft. At decoder scale $s$, the dominant operations are channel-wise feature normalization and same-location cross-correlation $O(H_sW_sC_s)$, followed by a lightweight $1\times1$ gate over $T_s=\min(\texttt{topk},C_s)$ channels and element-wise residual modulation $O(H_sW_sT_s)$. Because the module is attached only at selected decoder scales, it leaves the baseline backbone and prediction head unchanged.

A representative cross-model benchmark extends the previous single-model overhead view to include both baseline and \cosa{}-augmented variants where available. In the shared RTX A6000 batch-1 $256\times256$ forward-pass benchmark, FC-Siam+\cosa{} adds only 66 trainable parameters overall and changes GFLOPs only at very fine precision, while improving changed-class \Fonec{} by +1.75 points. STANet shows the clearest transfer gain (\Fonec{} 80.71$\rightarrow$86.14) with modest extra cost, whereas BIT changes only marginally in both accuracy and resource use. These comparisons reinforce the broader pattern from the hero-model section: \cosa{} is most useful when it complements weaker temporal interaction, and its overhead remains small relative to the affected backbone.

Table~\ref{tab:efficiency_overhead} is intentionally scoped as a common hardware benchmark rather than a unified accuracy ranking, because the compared backbones were trained under their preferred recipes and not all of them admit the same full-resolution evaluation path. In particular, STANet's native pairwise attention exceeded A6000 memory at full-resolution $1024\times1024$, so the shared cross-model benchmark is reported at $256\times256$. The batch size of 1 used in Table~\ref{tab:efficiency_overhead} is therefore an inference-profiling choice, not the training configuration: the controlled FC-Siam accuracy experiments still use batch size 8. To complement that shared benchmark with a deployment-scale view, our controlled FC-Siam full-resolution measurement on LEVIR-CD raises latency from 299.05 to 303.26 ms per image and peak GPU memory from 3396.83 to 3531.07 MB on the same GPU (3.34 to 3.30 FPS, batch size 1, input size $1024\times1024$). Taken together, these measurements make the performance/complexity tradeoff more concrete: \cosa{} introduces little architectural overhead, and its practical cost is small relative to the accuracy gain in the main controlled setting.

\section{Conclusion}\label{sec:conclusion}
This work introduced Context Sampling Attention (\cosa{}), a lightweight decoder-side refinement module for remote sensing change detection. In the implemented formulation studied here, \cosa{} uses pointwise bi-temporal feature correlation as an explicit control signal, converts low correlation into a learned change gate, and injects the gated residual at selected decoder scales through learnable residual scaling.

On four benchmarks (LEVIR-CD, S2Looking, DSIFN, and CLCD), \cosa{} improves changed-class F1 and IoU over the controlled baseline. Ablations indicate that multiscale placement and adaptive residual gating are both important for peak performance. Cross-backbone tests further show that the module is most helpful when temporal fusion is limited and can be near-neutral when fusion is already strong.

Despite these gains, limitations remain. \cosa{} is not uniformly beneficial across architectures, and some samples still degrade in ambiguous or noisy-boundary regions where correlation is less reliable. Moreover, although the controlled FC-Siam benchmark in this paper now includes measured latency and GPU-memory overhead, broader cross-hardware deployment profiling remains limited. Table~\ref{tab:efficiency_overhead} makes clear that the current deployment evidence is still tied to a shared single-hardware benchmark rather than a broader hardware sweep.

These limitations suggest several practical directions for future work. Confidence-aware gating could reduce unnecessary refinement on difficult samples, broader transfer studies could clarify more precisely where \cosa{} is complementary versus redundant, and larger standardized efficiency benchmarks could further strengthen the deployment case for lightweight correlation-conditioned refinement.

\begingroup
\sloppy

\endgroup

\end{document}